\documentclass[11pt]{article}

\usepackage{amsmath}
\usepackage{amssymb}
\usepackage{bm}
\usepackage{bbm}
\usepackage{natbib}
\usepackage{pgfplots}
\usepackage{xcolor}
\pgfplotsset{compat=1.17}
\usepgfplotslibrary{groupplots}
\usepackage{csvsimple}
\usepackage{booktabs}
\usepackage{makecell}
\usepackage{graphicx}
\usepackage{array}
\usepackage{enumerate}
\usepackage{todonotes}
\usepackage[left=3cm, right = 3cm, bottom = 2cm, top = 2cm]{geometry}
\usepackage{algorithm}
\usepackage{enumitem}
\usepackage[breaklinks=true]{hyperref}
\newcommand{\bx}{\bm{x}}
\newcommand{\bX}{\bm{X}}
\newcommand{\bz}{\bm{z}}
\newcommand{\bZ}{\bm{Z}}
\newcommand{\bbeta}{\bm{\beta}}
\newcommand{\btheta}{\bm{\theta}}

\newcommand{\Ex}{\operatorname{E}}

\DeclareMathOperator*{\argmin}{arg\,min}

\newtheorem{remark}{Remark}

\definecolor{LightGray}{RGB}{211, 211, 211}
\definecolor{DarkGray}{RGB}{64, 64, 64}
\definecolor{PleasantBlue}{RGB}{0, 123, 255}
\definecolor{FreshGreen}{RGB}{40, 167, 69}
\definecolor{WarmOrange}{RGB}{255, 133, 27}
\definecolor{DeepPurple}{RGB}{102, 51, 153}
\definecolor{BrightRed}{RGB}{220, 53, 69}
\definecolor{ClearCyan}{RGB}{23, 162, 184}

\author{
    Henning Zakrisson\thanks{Department of Mathematics, Stockholm University}{ }\thanks{Corresponding author: \texttt{zakrisson@math.su.se}} \and
    Mathias Lindholm\footnotemark[1]
}

\title{A tree-based varying coefficient model}

\begin{document}

    \maketitle

    \begin{abstract}
        \noindent The paper introduces a tree-based varying coefficient model (VCM) where the varying coefficients are modelled using 
the cyclic gradient boosting machine (CGBM) from~\cite{delong2023note}.
Modelling the coefficient functions using a CGBM allows for dimension-wise early stopping and feature importance scores.
The dimension-wise early stopping not only reduces the risk of dimension-specific overfitting, 
but also reveals differences in model complexity across dimensions.  
The use of feature importance scores allows for simple feature selection and easy model interpretation.
The model is evaluated on the same simulated and real data examples as those used
in~\cite{richman2023localglmnet}, and the results show that it produces results in terms of out of sample
loss that are comparable to those of their neural network-based VCM called LocalGLMnet.
\vspace{3mm}


    \end{abstract}

    \section{Introduction}
    \label{sec:introduction}
    Many machine learning models that yield accurate results and high predictive power struggle with another important factor -- interpretability.
This is especially a problem for deep learning models, see e.g.\ \cite{marcinkevivcs2023interpretable}.
One way to address this is to use a model structure that is inherently interpretable, such as a generalised linear model (GLM).
However, without manual feature engineering, GLMs can not handle non-linear effects and interactions between features.
A more flexible model family that retains some of the interpretability of GLMs are
varying-coefficient models (VCMs), introduced in~\cite{hastie1993varying}.
In a VCM, the regression coefficients are replaced by feature-dependent regression coefficient functions, which can
be of any, possibly non-linear, functional form.
A recent example of a VCM is the LocalGLMnet model introduced in~\cite{richman2023localglmnet},
where the regression coefficient functions are modeled using a feed-forward neural network (FFNN).
This creates a \textit{locally} interpretable machine learning model, as those discussed
in e.g.~\cite[Section 4.2.6]{marcinkevivcs2023interpretable}.
The LocalGLMnet shows promising accuracy on both simulated and real-world data,
while also allowing for more insightful feature selection and local interpretability. 
Other examples of VCMs combined with neural networks are
e.g.~\cite{alvarez2018towards, al2020contextual, thompson2023contextual}, 
and these are sometimes referred to as ``contextual'' models.

As an alternative to using feed-forward neural network (FFNN)-based models
for the coefficient functions, tree-based models are
generally easier to interpret, and they are often easier to tune and faster to train. 
In~\cite{wang2014boosted}, this idea is introduced, focusing on linear models and gradient boosting machines (GBMs) using vector-valued
trees, i.e.\ trees with response of the same dimension as the vector of predictive features.
Although the focus is on linear models, generalised linear models are also discussed. 
This ansatz, however, differs from applying a multidimensional GBM directly to the loss function induced by the VCM formulation.
On the other hand, the original GBM algorithm~\citep{friedman2001greedy} and most of its successors
such as XGBoost~\citep{chen2015xgboost} and LightGBM~\citep{ke2017lightgbm}
provide \textit{univariate} estimators, meaning that the output of the models is
one-dimensional.
While these models are useful for estimating e.g.\ the mean parameter of a
distribution, these models do not allow for fitting a local GLM structure with
coefficient functions depending on two or more features.
Recently, a number of \textit{multivariate} GBM models have been proposed, such
as Delta Boosting~\citep{lee2018delta}, NGBoost~\citep{duan2020ngboost} and Cyclic
Gradient Boosting (CGBM)~\citep{delong2023note}. 
In~\cite{zhou2022decision}, a GBM-based VCM is introduced where univariate GBMs are fitted to the partial derivatives
simultaneously in each boosting iteration.

The current paper introduces a CGBM-based VCM.
The main reason for using cyclic GBMs instead of the simultaneous updates using univariate GBMs as in~\cite{zhou2022decision} is
that this allows for dimension-wise early stopping, something not covered in~\cite{wang2014boosted, zhou2022decision}.
By using dimension-wise early stopping, it is possible to reduce the risk of dimension-wise overfitting of the coefficient functions.
Further, this also provides additional insights into differences in model complexity across dimensions.
Moreover, the cyclic updates allow for the use of dimension-wise feature importance scores.
These feature importance scores can be used both for model interpretation and removing redundant features from the model, 
reducing model complexity.

The main contribution of this paper is to use \textit{cyclic} GBMs as a way to model the varying coefficient functions in a VCM.
The framework is also closely related to the LocalGLM model from~\cite{richman2023localglmnet},
and the cyclically boosted VCM retrieves a tree-based version of the LocalGLMnet model as a special case.
The algorithm focuses on interpretability and ease of use -- by focusing the modeling effort on the coefficient functions and not allowing
for structure in the intercept term $\beta_0$.
Also, the CGBM algorithm allows for dimension-wise early stopping, which can be used to reduce the risk of overfitting, something
that is not possible in the LocalGLMnet model, nor the boosted VCM from~\cite{zhou2022decision}.
The dimension-wise feature importance score calculations of the CGBM algorithm also allow for easier model tuning and interpretation
of interactions between features.

The remainder of the paper is organised as follows:
Section~\ref{sec:model_architecture} introduces the VCM structure and the CGBM used to construct a cyclically boosted tree VCM.
Section~\ref{sec:modeling_considerations} discusses modeling considerations when training and using this model.
Section~\ref{sec:examples} presents examples of using cyclically boosted VCMs on simulated and real-world data, using 
the same data sets as in~\cite{richman2023localglmnet}.
The paper ends with concluding remarks in Section~\ref{sec:conclusion}.

    \section{Model architecture}
    Before going into varying coefficient models and local generalised linear models, we briefly introduce the concept of generalised linear models.
    \label{sec:model_architecture}

        \subsection{Generalised linear models}
        \label{subsec:glm}
        Let $Y\in \mathbb{R}$ denote the response, e.g.\ number of claims or claim cost,
and let $\bX \in \mathbb{X} \subseteq \mathbb{R}^p$ be a feature vector.
If we consider exponential dispersion family (EDF) models, 
see e.g.~\cite{jorgensen1997theory}
 and generalised linear models (GLMs),
see e.g.\ \cite{nelder1972generalized}, we assume that
\begin{align}\label{eq:glm_mean}
	\Ex[Y \mid \bX = \bx] = \mu(\bx; \beta_0,\bbeta) := u^{-1}\left(
\beta_0 + \bbeta^\top \bx
\right),
\end{align}
where $\bbeta\in\mathbb{R}^{p}$ and $u$ is the so-called link function.
When using EDF models there may be an additional dispersion parameter $\varphi$ that
is possible to neglect when estimating $\mu$ or $\bbeta$.
That is, neither the so-called deviance function nor the log-likelihood function
w.r.t.\ $\mu$ of an EDF will depend on dispersion parameters.
Moreover, in many actuarial applications, it is common to include
weights, $W$, corresponding to e.g.\ policy duration.
In order to ease the exposition, any potential dependence on weights will be
suppressed until the numerical illustrations in Section~\ref{sec:examples}.
For a longer discussion of duration effects,
see~\cite{lindholm2023local} and~\cite{lindholm2023duration}.

When it comes to estimation of model parameters in a GLM,
given an independent sample $(\bx_i, y_i)_{i = 1}^n$, this can be expressed as
\begin{equation}
    \label{eq:glm_mle}
    (\widehat{\beta}_0,\widehat{\bbeta}) = \argmin_{
        (\beta_0,\bbeta) \in \mathbb{R}^{p+1}
    } \left\{ \sum_{i=1}^n \mathcal{L}\left(\mu(\bx_i;
    \beta_0,\bbeta);
    y_i
\right)\right\},
\end{equation}
where $\mathcal{L}(\mu; y)$ corresponds to either the negative log-likelihood or the
deviance function for an observation $y$ given
mean parameter $\mu$~\citep{sundberg2019statistical}.
Further, as seen from~\eqref{eq:glm_mle} the general optimisation problem does
not explicitly depend on the link-transformed linear function from~\eqref{eq:glm_mean}.
Continuing, \eqref{eq:glm_mle} can be solved using gradient descent
where the partial derivatives
\begin{align}\label{eq:directional_partial_derivative}
	g_{ij} := &~ \frac{\partial}{\partial \beta_j}
\mathcal{L}(\mu(\bx_i; \beta_0,\bbeta); y_i)\nonumber\\
		= &~ \frac{\partial}{\partial \mu} \mathcal{L}(\mu; y_i)
\bigg|_{\mu = \mu(\bx_i; \beta_0,\bbeta)} \cdot \frac{\partial}{\partial \beta_j}
\mu(\bx_i; \beta_0,\bbeta)\nonumber\\
		= &~ x_{ij} \frac{\partial}{\partial \mu} \mathcal{L}(\mu; y_i)
\bigg|_{\mu = \mu(\bx_i; \beta_0,\bbeta)} \cdot \frac{\partial}{\partial v} u^{-1}(v)
\bigg|_{
	v = \beta_0 + \bbeta^\top \bx_i
	} 
\end{align}
appear.
Note that~\eqref{eq:directional_partial_derivative} can be considered as
a directional partial derivative. 
Further, note that if we use a log-link, i.e.
\begin{align}
	\mu(\bx; \beta_0,\bbeta) := \exp\left\{
	\beta_0 + \bbeta^\top \bx
	\right\},
\end{align}
then \eqref{eq:directional_partial_derivative} reduces to
\begin{equation*}
	g_{ij} :=~ x_{ij} \mu(\bx_i;\beta_0, \bbeta)\frac{\partial}{\partial \mu} 
\mathcal{L}(\mu; y_i)\bigg|_{\mu = \mu(\bx_i; \beta_0,\bbeta)}.
\end{equation*}
\begin{remark}\label{remark:GLM_loss_partial_derivatives}
	Note that the loss function $\mathcal{L}(\mu; y)$ has a parameter dimension 
of 1, given by $\mu \in \mathbb{R}$, whereas the parametrisation of the mean 
function $\mu(\bx;\beta_0, \bbeta)$ has $p+1$ dimensions
since $\bbeta \in \mathbb{R}^{p}$.
Further, for a fixed observation $i$, the partial derivatives
in~\eqref{eq:directional_partial_derivative} for different dimensions $j$
will be dependent.
\end{remark}

        \subsection{VCMs and local GLMs}
        \label{subsec:local_glm}
        The link-transformed linear relationship between the features in a GLM provides
models that are interpretable and numerically stable.
The GLMs, however, are in general not able to compete with machine learning models such
as gradient boosting machines (GBMs) and feed-forward neural networks
(FFNNs) in terms of predictive accuracy.
This is often the case even if extensive manual feature engineering has been
conducted.

The varying-coefficient model (VCM) framework, introduced in~\cite{hastie1993varying}, suggests that the
regression coefficients $\beta_0$, $\bbeta$ in the linear predictor of a GLM (see~\eqref{eq:glm_mean}) are replaced by
feature dependent regression coefficient functions $\beta_0(\bz)$, $\bbeta(\bz)$, where $\bz \in \mathbb{Z}$.
In~\cite{richman2023localglmnet} it is suggested to replace the regression coefficients, $\bbeta$, in \eqref{eq:glm_mean} to
regression coefficient functions $\bbeta(\bx)$, whereas $\beta_0$ is kept constant.
To generealize these propositions, we suggest to replace the regression coefficients, $\bbeta$, in \eqref{eq:glm_mean} to
regression coefficient functions $\bbeta(\bz)$, where $\bz \in \mathbb{Z}$, whereas $\beta_0$ is kept constant.
This yields the mean function
\begin{equation}
	\label{eq:local_glm_structure}
	\mu(\bx;\beta_0,\bbeta(\bz)) := u^{-1}(
		\beta_0 + \bbeta(\bz)^\top \bx
	).
\end{equation}
Note that it may be the case that $\mathbb{Z} \not\subset \mathbb{X}$.
For example, $\mathbb{Z}$ may include time as a feature, whereas $\mathbb{X}$ does not.

Note that keeping $\beta_0$ as a constant intercept, independent of $\bz$, is
not the case in a standard VCM, but chosen here in order for as much of the functional structure
as possible to be captured by the coefficient functions $\bbeta(\bz)$.
By analysing $\beta_j(\bz)x_j$, this also allows us to assess
potential interactions between $x_j$ and $\bz$.
If $\mathbb{X}_j \subset \mathbb{Z}$, where $\mathbb{X}_j$ is the $j$th feature space,
then $\beta_j(\bz)$ will reveal non-linear dependence on $\bx$.
In Appendix~\ref{app:l2_calculations} estimation of regression coefficient functions is discussed with a focus on the $L^2$ situation 
and population-based quantities.
It is there seen that the $j$th regression coefficient function can be expressed as a certain
expected value conditional on $\bZ_j$.
It is also seen that if the true mean function is of the form of a VCM, then the regression coefficient function estimators will 
retrieve, possibly noisy, versions of the corresponding true regression coefficient functions.
In particular, if in addition the true regression coefficient functions are constants, these constants will be estimated correctly,
as should be the case.

For more on interpretations and consequences of using~\eqref{eq:local_glm_structure}, 
see Remark~2.3 in~\cite{richman2023localglmnet}, and, more generally, see Section~2 in~\cite{hastie1993varying}.
We also elaborate on these topics below in the numerical illustrations in Section~\ref{sec:examples}.

Continuing, in~\cite{richman2023localglmnet} the coefficient functions are modeled
using feed-forward NNs.
In the present paper, we instead model the coefficient functions using GBMs,
but as commented on in Remark~\ref{remark:GLM_loss_partial_derivatives} the
loss w.r.t.\ $\bbeta$ has a parameter dimension of $p+1$.
Thus, to implement a GBM-based VCM in accordance 
with~\eqref{eq:local_glm_structure} calls for using a multi-dimensional GBM 
applied to the directional derivatives from~\eqref{eq:directional_partial_derivative}.
One way of doing this is to use the cyclic GBM (CGBM) from~\cite{delong2023note}.

            \subsubsection{The Cyclic Gradient Boosting Machine}
            \label{subsubsec:cgbm}
            The cyclic gradient boosting machine (CGBM) is introduced in~\cite{delong2023note},
and is concerned with estimating an unknown $d$-dimensional parameter function
$\btheta: \mathbb{X} \rightarrow \mathbb{R}^d$ using regression tress.
The CGBM, in its basic formulation, is a multi-parametric generalisation of the
standard GBM from~\cite{friedman2001greedy}.
The algorithm is presented in Algorithm~\ref{alg:cgbm}.
For further details, see~\cite{delong2023note}.
\begin{algorithm}[H]
	\caption{Cyclic Gradient Boosting Machine}
	\label{alg:cgbm}
	\textbf{Let:}
\vspace{-0.75em}
\begin{itemize}[itemsep=0pt,parsep=0pt]
    \item[$\bullet$] $\mathcal{L}( \btheta; y)$ be a loss function,
    \item[$\bullet$] $(y_i, \bx_i)_{i = 1}^n$ be independent observations,
    \item[$\bullet$] $\mathcal{A}$ be a disjoint partition of $\mathbb{X}$ into $|\mathcal{A}|$ regions,
    \item[$\bullet$] $\gamma_l \in \mathbb{R}$ be terminal node values, $l = 1, \ldots, |\mathcal{A}|$,
    \item[$\bullet$] $\epsilon_j$ and $\kappa_j$ be shrinkages and number of trees for parameter dimension $j$,
    \item[$\bullet$] $\boldsymbol{e}_j$ be the $j$th unit vector.
\end{itemize}
\vspace{-0.75em}
\textbf{Initiate} with $k=0$,
    \begin{equation}
        \label{eq:cgbm_init}
        \widehat{\btheta}(\bx) = \widehat{\btheta}^{(0)} :=  \argmin_{\btheta \in \mathbb{R}^d}
        \sum_{i = 1}^n \mathcal{L}(\btheta; y_i).
    \end{equation}
\vspace{-0.75em}
\textbf{While} $k < \max\limits_{j = 1, \ldots, d} \kappa_j$:
\vspace{0.5em}
\begin{itemize}[itemsep=0pt,parsep=0pt]
    \item[] $k \leftarrow k + 1$
    \item[] \textbf{For} $j = 1, \ldots, d$:
    \begin{itemize}[itemsep=0pt,parsep=0pt]
        \item[] \textbf{If} $k \leq \kappa_j$:
        \begin{enumerate}[itemsep=0pt,parsep=0pt,label=(\roman*)]
            \item \textbf{Calculate} partial derivatives, for $i = 1, \ldots, n$:
                \begin{equation}
                    \label{eq:cgbm_partial_derivatives}
                    g_{ij} = \frac{\partial}{\partial \theta_j}
                    \mathcal{L}(\btheta; y_i)\bigg|_{\btheta = \widehat{\btheta}(\bx_i)},
                \end{equation}
            \item \textbf{Fit} tree regions
            \[ \widehat{\mathcal{A}}^{(k,j)} = \argmin_{
                \mathcal{A} }
                \sum_{\mathcal{A}_l \in \mathcal{A}}
                \sum_{i: \bx_i \in \mathcal{A}_l}
                \left(
                    g_{ij} - \bar{g}_l
            \right)^2
            \]
            where $\bar{g}_l$ is the average of $g_{ij}$ for observations
            s.t.\ $\bx_i \in \mathcal{A}_l$.
            \item \textbf{Adjust} terminal node values
            \[
                \widehat{\gamma}^{(k,j)}_l = \argmin_{\gamma \in \mathbb{R}}\
                \sum_{i : \bx_i \in \widehat{\mathcal{A}}_{l}^{(k,j)}}
                {\mathcal{L}}(
                \widehat{\btheta}(\bx_i) + \gamma \epsilon_j \boldsymbol{e}_j, y_i), \quad
                l = 1, \ldots, |\widehat{\mathcal{A}}^{(k,j)}|.
            \]
            \item \textbf{Update} parameter function
            \[
                \widehat {\btheta}(\bx) \leftarrow  \widehat {\btheta}(\bx)
                + {\boldsymbol{e}}_j \epsilon_{j}
                \sum_{\mathcal{A}_l \in \widehat{\mathcal{A}}^{(k,j)}}
                \mathbbm{1}_{\{\bx \in \mathcal{A}_l\}}
                \widehat{\gamma}^{(k,j)}_l.
            \]
        \end{enumerate}
    \end{itemize}
\end{itemize}
\vspace{-0.75em}
\textbf{Return} 
\[
    \widehat{\btheta}(\bx) = \widehat{\btheta}^{(0)} + 
    \sum_{j = 1}^d
    \sum_{k = 1}^{\kappa_j}
    \sum_{\mathcal{A}_l \in \widehat{\mathcal{A}}^{(k,j)}}
    \boldsymbol{e}_j
    \epsilon_j
    \mathbbm{1}_{\{\bx \in \mathcal{A}_l\}}
    \widehat{\gamma}^{(k,j)}_l
\]

\end{algorithm}

Note that the individual components of $\widehat{\btheta}(\bx)$ can be
written
as individual GBMs, i.e.
\begin{equation*}
\widehat{\theta}_j(\bx)
=
\widehat{\theta}^{(0)}_j
+ \sum_{k = 1}^{\kappa_j}
\sum_{\mathcal{A}_l \in \widehat{\mathcal{A}}^{(k,j)}}
\epsilon_j
\widehat{\gamma}^{(k,j)}_l
\end{equation*}
The standard, univariate, GBM from~\cite{friedman2001greedy} is
obtained by setting $d = 1$.

Since GBMs are overparameterised, the CGBM is prone to overfitting.
To regularise the CGBM, one needs to find appropriate hyperparameters (e.g.\ max depth and 
minimum samples per terminal region) for the trees,
reasonable shrinkage factors $\epsilon_j$ and number of trees $\kappa_j$.
Algorithm 3 in~\cite{delong2023note} describes a procedure for finding appropriate
values of $\kappa_j$ using early stopping, given a fixed value of $\epsilon_j$.
The algorithm examines loss function contributions when adding new trees,
and is based on the idea that trees should be added to dimension $j$ of the
parameter function until an additional tree no longer decreases the loss.
The method utilises a validation data set or cross-validation to determine when
to stop adding trees.
This means that if there is no evidence of any structure in the parameter 
dimension $j$, i.e. $\theta_j(\bx) \equiv \theta^{(0)}_j$,
the algorithm will, given a sufficient amount of data, set $\kappa_j = 0$.

As mentioned earlier, GBMs are generally considered to be more interpretable
than FFNNs.
One reason for this is the ability to calculate
\textit{feature importance scores} as defined by~\cite{breiman1984classification}.
The feature importance score $\text{FI}_j$ of a feature $x_j$ in a regression tree is defined
as the total reduction of the loss function when using feature $x_j$ in any of
the binary splits in the tree partition $\mathcal{A}$ of the feature space.
\cite{friedman2001greedy} uses the same score for the GBM, and proposes that the
feature importance scores can be calculated for the GBM by summing the
feature importance scores for each tree in the model.
This property is inherited by the CGBM and can be used to assess the
importance of features not only for the overall model -- but also for each
individual component $\theta_j(\bx)$.
This is done by calculating the total reduction of the loss function when using
feature $x_k$ in any of the trees that are used in the boosting steps of
parameter dimension $j$.

            \subsubsection{A cyclically boosted VCM}
            \label{subsubsec:localglmboost}
            As discussed in the beginning of Section~\ref{subsec:local_glm},
the local GLM formulation~\eqref{eq:local_glm_structure} can be estimated using
the CGBM using $\btheta(\bz) = \bbeta(\bz)$ based on the loss function $\mathcal{L}(\mu(\bx; \beta_0,\bbeta);y)$,
where $\bbeta \in  \mathbb{R}^{p}$. 
This leads to the observation-specific directional partial derivatives
\[
	g_{ij} = x_{ij} \frac{\partial}{\partial \mu} \mathcal{L}(
\mu
; y_i)\bigg|_{
	\mu = \mu(\bx_i; \beta_0,\bbeta)
	}
\cdot \frac{\partial}{\partial v} u^{-1}(v)
\bigg|_{
	v = \beta_0 + \bbeta^\top \bx_i
	}
\]
from~\eqref{eq:directional_partial_derivative}.
That is,~\eqref{eq:cgbm_partial_derivatives}
can be written as
\begin{equation}
	g_{ij} = x_{ij} \frac{\partial}{\partial \mu} \mathcal{L}(
\mu
; y_i)\bigg|_{
	\mu = \mu(\bx_i; \widehat{\beta}_0, \widehat{\bbeta}(\bz_i))
	}
\cdot \frac{\partial}{\partial v} u^{-1}(v)
\bigg|_{
	v = \widehat{\beta}_0 + \widehat{\bbeta}(\bz_i)^\top \bx_i
	}.
\end{equation}
By using the above directional partial derivatives in the CGBM from
Section~\ref{subsubsec:cgbm} we obtain a cyclically boosted
VCM.
The natural initialisation of the model is 
using~\eqref{eq:cgbm_init}, which is equivalent to initialising with a standard GLM, see~\eqref{eq:glm_mle}.
See Algorithm~\ref{alg:localglmboost} for the full training procedure.
\begin{algorithm}[H]
	\caption{A cyclically boosted VCM}
	\label{alg:localglmboost}
	\textbf{Let:}
\vspace{-0.75em}
\begin{itemize}[itemsep=0pt,parsep=0pt]
    \item[$\bullet$] $\mathcal{L}(\mu; y)$ be a loss function,
    \item[$\bullet$] $(y_i, \bx_i, \bz_i)_{i = 1}^n$ be independent observations,
    \item[$\bullet$] $\mu(\bx; \beta_0,\bbeta) = u^{-1}(\beta_0 + \bbeta^\top \bx)$ be a mean function,
    \item[$\bullet$] $\mathcal{A}$ be a disjoint partition of $\mathbb{Z}$ into $|\mathcal{A}|$ regions,
    \item[$\bullet$] $\gamma_l \in \mathbb{R}$ be terminal node values, $l = 1, \ldots, |\mathcal{A}|$,
    \item[$\bullet$] $\epsilon_j$ and $\kappa_j$ be shrinkage parameters and number of trees for coefficient function $j$,
\end{itemize}
\vspace{-0.75em}
\textbf{Initiate} with $k=0$,
    \begin{equation}
        \label{eq:localglmboost_init}
        (\widehat{\beta}_0,\widehat{\bbeta}(\bz)) =
        (\widehat{\beta}_0,\widehat{\bbeta}^{\operatorname{GLM}})
        :=  \argmin_{(\beta_0,\bbeta)\in \mathbb{R}^{p+1}}
        \sum_{i = 1}^n \mathcal{L}(\mu(\bx_i; \beta_0, \bbeta), y_i).
    \end{equation}
\vspace{-0.75em}
\textbf{While} $k < \max\limits_{j = 1, \ldots, p} \kappa_j$:
\begin{itemize}[itemsep=0pt,parsep=0pt]
    \item[] $k \leftarrow k + 1$
    \item[] \textbf{For} $j = 1, \ldots, p$:
    \begin{itemize}[itemsep=0pt,parsep=0pt]
        \item[] \textbf{If} $k \leq \kappa_j$:
        \begin{enumerate}[itemsep=0pt,parsep=0pt,label=(\roman*)]
            \item \textbf{Calculate} partial derivatives, for $i = 1, \ldots, n$:
                \begin{equation*}
                    g_{ij} = x_{ij} \frac{\partial}{\partial \mu} \mathcal{L}(
                        \mu
                        ; y_i)\bigg|_{
                            \mu = \mu(\bx_i; \widehat{\beta}_0, \widehat{\bbeta}(\bz_i))
                            }
                        \cdot \frac{\partial}{\partial z} u^{-1}(v)
                        \bigg|_{
                            v = \widehat{\beta}_0 + \widehat{\bbeta}(\bz_i)^\top \bx_i
                            }.
                \end{equation*}
            \item \textbf{Fit} tree regions
            \begin{equation} \widehat{\mathcal{A}}^{(k,j)} = \argmin_{
                \mathcal{A} }
                \sum_{\mathcal{A}_l \in \mathcal{A}}
                \sum_{i: \bz_i \in \mathcal{A}_l}
                \left(
                    g_{ij}^{(k)} - \bar{g}_l^{(k)}
            \right)^2
            \label{eq:localglmboost_fit}
            \end{equation}
            where $\bar{g}_l^{(k)}$ is the average of $g_{ij}^{(k)}$ for observations
            s.t.\ $\bz_i \in \mathcal{A}_l$.
            \item \textbf{Adjust} terminal node values, for $l = 1, \ldots, |\widehat{\mathcal{A}}^{(k,j)}|$:
            \[
                \widehat{\gamma}^{(k,j)}_l = \argmin_{\gamma \in \mathbb{R}}\
                \sum_{i : \bz_i \in \widehat{\mathcal{A}}_{l}^{(k,j)}}
                {\mathcal{L}}\left(
                u^{-1}\left( \widehat{\beta}_0 + \widehat{\bbeta}(\bz_i)^\top \bx_i
                +
                \gamma x_{ij}  \right), y_i\right).
            \]
            \item \textbf{Update} coefficient function
            \[
                \widehat {\beta}_j(\bz) \leftarrow  \widehat {\beta}_j(\bz)
                + \epsilon_{j}
                \sum_{\mathcal{A}_l \in \widehat{\mathcal{A}}^{(k,j)}}
                \mathbbm{1}_{\{\bz \in \mathcal{A}_l\}}
                \widehat{\gamma}^{(k,j)}_l.
            \]
        \end{enumerate}
    \end{itemize}
\end{itemize}
\textbf{Restore unbiasedness} 
    \begin{equation}
    \label{eq:localglmboost_unbiasedness}
    \widehat{\beta}_0 = \argmin_{\beta_0 \in \mathbb{R}}
    \sum_{i = 1}^n \mathcal{L}(u^{-1}(\beta_0 + \widehat{\bbeta}(\bz_i)^\top \bx_i); y_i).
    \end{equation}

\vspace{-0.75em}
\textbf{Return} 
\begin{align}
    \widehat{\mu}(\bx) &= u^{-1}( \widehat{\beta}_0 + \sum_{j = 1}^p \widehat{\beta}_j(\bz) x_j) \nonumber\\
    \widehat{\beta}_j(\bz) &= \widehat{\beta}_j^{\operatorname{GLM}} +
    \epsilon_j
    \sum_{k = 1}^{\kappa_j}
    \sum_{\mathcal{A}_l \in \widehat{\mathcal{A}}^{(k,j)}}
    \mathbbm{1}_{\{\bz \in \mathcal{A}_l\}}
    \widehat{\gamma}^{(k,j)}_l, \quad j = 1, \ldots, p.
    \label{eq:localglmboost_return}
\end{align}

\end{algorithm}
Note that the intercept term $\beta_0$ is not updated during the boosting
iterations, and will be kept fixed at the initial value from~\eqref{eq:localglmboost_init}.
This is done to ensure the structure of the data is instead captured by the regression
coefficient functions.

Using the initialisation in~\eqref{eq:localglmboost_init} it is clear that this model
is a generalisation of a standard GLM, and that we may express the tree-based coefficient function estimates
in~\eqref{eq:localglmboost_return} as
\begin{align}
	\widehat\beta_j(\bz) =
	\widehat\beta_j^{\operatorname{GLM}} + \widehat\Delta_j(\bz),
\end{align}
where
\begin{align}
	\widehat\Delta_j(\bz) := \epsilon_{j}
	\sum_{k = 1}^{\kappa_j}
	\sum_{\mathcal{A}_l \in \widehat{\mathcal{A}}^{(k,j)}}
	\mathbbm{1}_{\{\bz \in \mathcal{A}_l\}}
	\widehat{\gamma}^{(k,j)}_l.
	\label{eq:localglmboost_delta}
\end{align}
By using this representation, it becomes straightforward to assess in which dimensions,
and parts of the effect modifier feature space that the model deviates
from the corresponding GLM.
Note also that if one uses a log-link, the estimate of the mean function
can be written as
\begin{align}
	\widehat{\mu}(\bx; \beta_0, \bbeta(\bz)) & = \exp\left\{\widehat{\beta}_0^{\operatorname{GLM}} +
	\sum_{j = 1}^{p} \widehat{\beta}_j^{\operatorname{GLM}} x_j +
	\sum_{j = 1}^{p} \widehat{\Delta}_j(\bz) x_j\right\} \\
	& = \widehat{\mu}^{\operatorname{GLM}}(\bx) \cdot \exp\left\{\sum_{j = 1}^{p}
	\widehat{\Delta}_j(\bz) x_j\right\}
\end{align}
meaning that a boosted VCM can be seen as a GLM with a
multiplicative correction term.

Once the model has been estimated, the balance property needs
to be restored to achieve global unbiasedness.
This is why the intercept term $\beta_0$ is updated in~\eqref{eq:localglmboost_unbiasedness}.

\begin{remark}
	Note that the regression coefficient functions $\beta_j(\bz)$ in the cyclically boosted VCM do not
	necessarily have to depend on the same effect modifiers $\bz$.
	It is straightforward to extend the model to allow for different
	effect modifiers for each coefficient function, i.e. replacing the mean function
	in~\eqref{eq:local_glm_structure} with
	\[
		\mu(\bx;\beta_0,\beta_1(\bz_1),\ldots,\beta_p(\bz_p)) := u^{-1}(
			\beta_0 + \sum_{j = 1}^p \beta_j(\bz_j) x_j
		).
	\]
	The only difference in Algorithm~\ref{alg:localglmboost} would be that the
	features used for splitting the trees in~\eqref{eq:localglmboost_fit} would be
	different for each coefficient function.
	\label{rem:localglmboost_diff_effect_modifiers}
\end{remark}

Note that the feature importance scores $\text{FI}_j$ discussed in Section~\ref{subsubsec:cgbm} can be calculated for 
the individual coefficient functions $\beta_j(\bz)$, in order to see which features
are important for each coefficient function.
\cite{wang2014boosted} use the feature importance score of the total model
to assess the importance of features for the individual coefficient functions.
This can be a useful measure if one assumes that the predictive features are standardised to be on the same scale --
and that $\mathbb{X} \cap \mathbb{Z} = \emptyset$.
However, if $\mathbb{X} \cap \mathbb{Z} \neq \emptyset$, the feature importance scores
for effect modifier $x_j$ for the full 
model is no longer as useful -- since the effect modifier $x_j$ will then also
effect the output by the term $\beta_j(\bz) x_j$, which will not be captured by the feature importance score.
In~\cite{richman2023localglmnet} the importance measure
\begin{equation}
	\mathrm{FI}_j^* =
	\frac{1}{n}
	\sum \limits_{i=1}^n
	\left| \widehat{\beta}_j(\bz_i)\right|
	\label{eq:variable_importance}
\end{equation}
is proposed for local GLMs, see Section 3.5 in~\cite{richman2023localglmnet}, where $\widehat{\beta}_j(\bz_i)$
is the coefficient function estimate $\widehat{\beta}_j(\bz)$ evaluated at
$\bz_i$ and $i=1,\ldots,n$ is the training sample.
The scores from~\eqref{eq:variable_importance} are normalised to sum to one and are used
to assess the importance of feature $x_j$ based on its regression coefficient function.
This is materially different from the feature importance scores used to assess the importance of feature $x_j$
\textit{within} a regression coefficient model.
Note that $\mathrm{FI}_j^*$, too, assumes that the features are standardised to be on the same scale,
and will neglect the potential effect of feature $x_j$ on other coefficient functions $\beta_k(\bz)$, $k \neq j$
in cases where $\mathbb{X} \cap \mathbb{Z} \neq \emptyset$.

            \subsubsection{Comments on convergence}
            \label{subsubsec:localglmboost_convergence}
            As discussed in Section~\ref{subsubsec:localglmboost}, the cyclically boosted VCM is constructed using the CGBM from~\cite{delong2023note}.
The CGBM aims at estimating an unknown $d$-dimensional parameter function $\btheta(\bx)$.
In the special case of the model, the unknown parameter function corresponds to the mean function 
from~\eqref{eq:local_glm_structure}, which is expressed in terms of $p + 1$ different components, 
i.e.\ the intercept $\beta_0$ together with the coefficient functions $\beta_i(\bx), i = 1, \ldots, p$.
Further, for EDFs, modeling an unknown, but in the local GLM and VCM sense, structured mean function, it is
natural to consider deviance loss-functions, see e.g.\ Chapter 1 in \cite{jorgensen1997theory}, given by
\begin{align}\label{eq: deviance loss}
	D(\mu; (y_i)_{i = 1}^n) = \sum_{i = 1}^n d(\mu; y_i),
\end{align}
where
\begin{equation*}
	d(\mu; y_i) \propto \log L(y_i; y_i) - \log L(\mu; y_i),
\end{equation*}
where $\log L(\mu; y)$ is the log-likelihood for a single observation w.r.t.\ an unknown mean parameter $\mu$ in an EDF.
Thus, by using $\mathcal{L}(\mu; y_i) = D(\mu;  y_i)$, it follows that the MLE minimises the deviance loss.
Moreover, from \eqref{eq: deviance loss}, it follows that the deviance loss-function attains its minimum, which is zero,
when the saturated model is reached, i.e.\ when $\widehat\mu(\bx_i) = y_i$.
Consequently, running the CGBM algorithm without early stopping will result in in-sample convergence,
see Proposition 1 in~\cite{delong2023note} and the surrounding discussion, since no cyclic update can result in an increasing loss.
In particular, if the partial derivative in a specific covariate dimension, say $j$, in iteration $k$, can not be improved,
the added tree will have terminal node values equal to 0. Similarly, if the partial derivative attains the value 0, the added tree 
will again have terminal node values equal to 0. This, however, does not guarantee that the minimum is unique in terms of the 
coefficient functions, it only tells us that the algorithm will, in-sample, converge to a local minimum, 
see Lemma 1 in~\cite{delong2023note}.

The above discussion only focuses on in-sample convergence. The question of out-of-sample (population-based) consistency is considerably harder to address.
In~\cite{zhou2022decision} consistency is discussed for the situation with $L^2$-boosting when the univariate
tree-approximations are updated jointly by making use of the construction from~\cite{buhlmann2002consistency}.
This approach, however, does not apply to the CGBM; see the discussion in Section 5 in~\cite{zhou2022decision}.
For more on the convergence of the CGBM, see the discussion in~\cite{delong2023note} relating to the results in~\cite{zhang2005boosting}.
These ideas are not pursued further in the current paper.

    \section{Modeling considerations}
    \label{sec:modeling_considerations}
    Three topics concerning modeling using VCMs, and in particular the tree-based VCM introduced in this paper,
will be discussed further, namely feature selection, interaction effects, and the
handling of categorical variables.

\begin{remark}
    The statements on the interpretation of coefficient functions in local GLMs,
    presented in Remark 2.3 of~\cite{richman2023localglmnet} apply to VCMs in general, and the model proposed in this paper in
    particular, and not only to local GLMs.
    We re-iterate them here for completeness, with some additional comments.
    \begin{enumerate}[label=(\roman*)]
        \item If $\beta_j(\bz) = \beta_j \neq 0$, i.e.\ the function is non-zero but does
              not depend on $\bz$, then the structure apart from the GLM term,
               e.g.~\eqref{eq:localglmboost_delta} in a VCM, is
               unnecessary.
        \item If $\beta_j(\bz) \equiv 0$, the term $\beta_j(\bz)x_j$ should not
              be included in the model. Note, however, that we can still use $x_j$ and $\bz$
              as input to other coefficient functions, i.e. having $\mathbb{X}_j \subset \mathbb{Z}$.
        \item If $\beta_j(\bz) =\beta_j(z_k)$, i.e.\ the coefficient function for feature $j$
              only depends on $z_k$, there are no effects from other features
                   any other feature for this term, and they can 
                  be removed as input to $\beta_j$ (see Remark~\ref{rem:localglmboost_diff_effect_modifiers}).
        \item Note that when $\mathbb{Z}= \mathbb{X}$, as in~\cite{richman2023localglmnet},
              we can never guarantee identifiability of the coefficient functions,
               since we can express any term $\beta_j(\bx)x_j$ as
              \[
                \beta_j(\bx)x_j = \beta_k(\bx)x_k
              \]
              using $\beta_k(\bx) = \beta_j(\bx)x_j/x_k$. 
              However, in the numerical illustration in Section~\ref{subsec:simulated_data},
              this does not seem to be a big problem since the "true" coefficient functions
              were estimated with reasonable accuracy. 
              See also Appendix~\ref{app:12_coefficient_function_interpretations} for a discussion
               on identifiability.
      \end{enumerate}
    \label{remark:regression_attentions}
\end{remark}

We continue with discussing specific properties of the CGBMs and their implications on the modeling of coefficient functions.

        \subsection{Feature selection}
        \label{subsec:feature_selection}
        Feature selection, i.e.\ picking predictive features $\bx$ (for picking effect modifier features $\bz$,
see Section~\ref{subsec:interaction_effects}), for the cyclically boosted VCM is
included automatically in the dimension-dependent early stopping procedure.
The early stopping scheme presented in~\cite{delong2023note} uses
different stopping times $\kappa_j$ for different parameter dimensions $j$.
In the cyclically boosted VCM, this means that a different number of trees can be
used for different coefficient function estimates $\widehat{\beta}_j(\bz)$.
The early stopping scheme is based on the idea that the model is trained until
the loss function is no longer reduced by adding new trees.
This is done for each dimension and for a feature $j$ where there is no evidence of coefficient function estimate
$\widehat{\beta}_j(\bx)$ providing any additional information over the initiation
$\widehat{\beta}_{j}^{\operatorname{GLM}}$, the hyperparameter $\kappa_j$ will
be close to zero. 
Also, if $\widehat{\beta}_{j}^{\operatorname{GLM}} = 0$, the feature $j$ will not be used in the model unless it
is included in the effect modifier set $\mathbb{Z}$.
Consequently, the early stopping procedure will automatically remove or at least reduce the impact of redundant features. 
This is a simplification compared to the method used for FNNs in~\cite{richman2023localglmnet}.

Further, by analysing the feature importance scores for each coefficient function (see Section~\ref{subsec:interaction_effects}),
 it is straightforward to manually reduce the model complexity 
by removing features that have little influence on a specific coefficient function.
In order to avoid drawing conclusions based on noise, this step could, e.g., be combined with a bootstrap procedure.

        \subsection{Interaction effects}
        \label{subsec:interaction_effects}
        The fact that it is possible to compute dimension-wise feature importance scores for the coefficient function estimates
in the cyclically boosted VCM makes it simple to assess interaction effects between effect modifiers $\bz$
and predictive features $\bx$.
This is done by examining the feature importance of effect modifier $z_k$ on coefficient function estimate
$\widehat{\beta}_j(\bz)$.
If a feature importance score
for effect modifier $z_k$ on coefficient function estimate $\widehat{\beta}_j(\bz)$ is $\approx 0$,
 this indicates no interaction effect between $z_k$ and $x_j$.
The effect modifier $z_k$ can then be removed from that coefficient function estimate, i.e.\
by removing $z_k$ from $\bz_j$ (see Remark~\ref{rem:localglmboost_diff_effect_modifiers}).
If an effect modifier gets no feature importance score for any coefficient function estimate, 
it can be removed from the effect modifier vector completely.
Note that it might still be included in the predictive feature vector $\bx$.
This step could be combined with a bootstrap procedure.
The above procedure is a simplification compared to the method used for FNNs in~\cite{richman2023localglmnet}.

        \subsection{Categorical features}
        \label{subsec:categorical_features}
        The VCM structure from \eqref{eq:local_glm_structure} allows for the use of categorical features
in the effect modifier vector $\bz$ as long as the chosen functional form of the coefficient functions $\beta_j(\bz)$
can handle categorical features.
Since the coefficient functions $\beta_j(\bz)$ are modelled as GBMs in the model proposed in this paper,
the model can handle categorical features natively.
Note, however, that in order to use categorical features in the predictive features $\bx$,
one has to encode them as one-hot encoded features, as done in~\cite{richman2023localglmnet}.
This means a categorical feature $x_j$ is replaced by a set of binary features $\bx^*_j$,
where each feature $\bx^*_{j,l}$ corresponds to a category $c_l$ of feature $x_j$,
i.e.
\begin{equation*}
    \bx^*_{j} = \begin{pmatrix}
        \mathbbm{1}_{\{x_j = c_1\}} \\
        \vdots \\
        \mathbbm{1}_{\{x_j = c_{m_j}\}} \\
    \end{pmatrix},
\end{equation*}
where $m_j$ is the number of categories of feature $x_j$.
Note that one usually removes one of the categories when using this encoding in
GLM settings since the intercept term $\beta_0$ is included in the model.
This will not be done here, since the coefficient functions
will be modeled as GBMs, whereas the intercept term $\beta_0$ will not.

Note that the update steps of a set of one-hot encoded predictive features can be
performed in parallel since the multiplicative term structure makes the updates of
coefficient function estimates $\widehat{\beta}_{j,l}(\bz)$ independent of each other for
all $l = 1, \ldots, m_j$ (see Algorithm~\ref{alg:localglmboost}).
This speeds up the estimation procedure, making it a practically implementable approach.

    \section{Examples}
    \label{sec:examples}
    The following numerical examples illustrate the performance of the proposed method on simulated and real data. 
Throughout this section we will use $\mathbb{Z} = \mathbb{X}$, and let $\mu(\bx) := \mu(\bx; \beta_0, \bbeta(\bx))$
Both datasets are the same as the ones used in Section 3 of~\cite{richman2023localglmnet}.
The \texttt{python} implementation of this special case of our VCM is available at
\url{https://github.com/henningzakrisson/local-glm-boost} under the name \texttt{LocalGLMboost}.

        \subsection{Simulated data}
        \label{subsec:simulated_data}
        The following simulated example reproduces the one in Section 3.1 of~\cite{richman2023localglmnet}.
The data is generated using features $\bX \in \mathbb{R}^8$, where
$\bX$ follows a multivariate normal distribution with mean vector $\boldsymbol 0$ and unit variances.
The features are independent, except for $X_2$ and $X_8$ which have a correlation of $0.5$.
The true regression function $\mu(\bx)$ is defined as
\begin{equation*}
    \mu\left(\bx\right)
    =
    \bbeta(\bx)^\top \bx
\end{equation*}
where $\bbeta(\bx) = (\beta_1(\bx), \dots, \beta_8(\bx))^\top$ is defined in
Table~\ref{tab:simulated_true_regression_attentions}.
\begin{table}[ht]
    \centering
    {
\renewcommand{\arraystretch}{1.25}
\begin{tabular}{c|l}
    Attention & Expression \\
    \midrule
    $\beta_1(\bx)$ & $0.5$  \\ 
    $\beta_2(\bx)$ & $-\frac{1}{4}x_2$  \\ 
    $\beta_3(\bx)$ & $\frac{1}{2} \textrm{sgn}(x_3)\sin(2x_3)$  \\ 
    $\beta_4(\bx)$ & $\frac{1}{4}x_5$  \\
    $\beta_5(\bx)$ & $\frac{1}{4}x_4$  \\ 
    $\beta_6(\bx)$ & $\frac{1}{8}x_5^2$  \\
    $\beta_7(\bx)$ & $0$ \\ 
    $\beta_8(\bx)$ & $0$ \\
    \bottomrule
\end{tabular}
}

    \caption{True coefficient functions $\beta_j(\bx)$ for the simulated data example.}
    \label{tab:simulated_true_regression_attentions}
\end{table}
Note that $X_7$ and $X_8$ are not included in the true regression function
as either effect modifiers or predictive features,
but $X_8$ is correlated with $X_2$.
Note also that the coefficient function representation is not unique for this $\mu(\bx)$
(see Remark~\ref{remark:regression_attentions} and Appendix~\ref{app:l2_numerical_example}).
Responses $y$ are generated according to
\[
    Y_i \mid \bX = \bx_i \sim \mathcal{N}\left(\mu\left(\bx_i\right), 1\right).
\]
In Appendix~\ref{app:l2_calculations} the $\beta_j(\bx)$s from Table~\ref{tab:simulated_true_regression_attentions}
are identified as the optimal solutions to a local GLM, or VCM, when knowing that the true
$\mu(\bx)$ is of the form \eqref{eq:local_glm_structure}.
Henceforth, the $\beta_j(\bx)$s from Table~\ref{tab:simulated_true_regression_attentions} will be referred to as the true
coefficient functions.

A total number of $200\,000$ observations are generated and split into a training and a test set of equal size.
The cyclically boosted tree-based VCM presented in Section~\ref{subsubsec:localglmboost}, henceforth referred to as the TVCM,
 is tuned using dimension-wise early stopping according to Algorithm 3
in~\cite{delong2023note}, using two equally sized training and validation splits. 
This produces a set of stopping times $\kappa_j$, $j = 1, \dots, 8$.
This algorithm relies on setting the other tree and boosting hyperparameters
to reasonable values without any in-depth evaluation; the max depth is set
to $2$, the minimum samples per leaf node is set to $10$, and the learning rate $\epsilon_j$ is
set to 0.01, $j = 1, \dots, 8$.

The results are compared to the true regression function $\mu(\bx)$ as well as
the results of the LocalGLMnet model from~\cite{richman2023localglmnet}.
An intercept-only model and a GLM without interactions are also included for comparison.
Since the negative log-likelihood of the normal distribution is proportional to the
Mean Squared Error (MSE), the MSE is used as the evaluation metric.

The MSE results for the training and test data can be seen in
Table~\ref{tab:simulated_data_mse}.
\begin{table}[ht]
    \centering
    \begin{tabular}{lrr}
{} & Train & Test \\
\midrule
\begin{filecontents*}{data/sim_loss.csv}
Model,Train,Test
True,1.002,0.996
Intercept,1.791,1.792
GLM,1.524,1.527
TVCM,1.008,1.013
LocalGLMnet,1.002,1.005
\end{filecontents*}
\csvreader[
 head to column names,
 late after line=\\
]{data/sim_loss.csv}{}%
{\Model & \Train & \Test}
\bottomrule
\end{tabular}

    \caption{MSE results for the simulated data example.}
    \label{tab:simulated_data_mse}
\end{table}
From Table~\ref{tab:simulated_data_mse} it is seen that the TVCM
is able to capture the true regression function well, with an MSE close
to the expected (being $1.00$).
It is also seen that an intercept-only model and a GLM without interactions perform 
poorly.
The performance of the TVCM is close to that of the LocalGLMnet.

\begin{figure}[ht]
    \centering
    \begin{tikzpicture}
    \begin{groupplot}[
        group style={
            group size=2 by 1,
            horizontal sep=2cm
        },
        width=6cm,
        height=6cm,
        xtick pos=left,
        ytick pos=left,
        tick style={major tick length=-1mm},
        xlabel = {$\mu(\bx)$},
        ylabel = {$\widehat{\mu}(\bx)$},
        set layers,
    ]
    
    \nextgroupplot[title=GLM]

        \addplot[
            samples=100,
            color=DarkGray,
            thick,
            on layer=axis foreground,
        ] table[
            x=mu,
            y=mu,
            col sep=comma,
        ] {data/sim_mu_hat.csv};

        \addplot[
            only marks,
            mark size=1pt,
            mark options={
                draw=FreshGreen,
                fill=FreshGreen,
                opacity=0.3
            },
            on layer=axis background
        ] table[
            x=mu,
            y=mu_GLM,
            col sep=comma,
        ] {data/sim_mu_hat.csv};

    \nextgroupplot[title=TVCM]

        \addplot[
            samples=100,
            color=DarkGray,
            thick,
            on layer=axis foreground,
        ] table[
            x=mu,
            y=mu,
            col sep=comma,
        ] {data/sim_mu_hat.csv};

        \addplot[
            only marks,
            mark size=1pt,
            mark options={
                draw=PleasantBlue,
                fill=PleasantBlue,
                opacity=0.3
            },
            on layer=axis background
        ] table[
            x=mu,
            y=mu_TVCM,
            col sep=comma,
        ] {data/sim_mu_hat.csv};

    \end{groupplot}

\end{tikzpicture}
    \caption{
        Estimated means $\widehat{\mu}(\bx)$ vs true regression function $\mu(\bx)$
        for $500$ randomly selected observations from the test set for the GLM and
        TVCMs respectively.
    }
    \label{fig:simulated_data_estimated_means}
\end{figure}
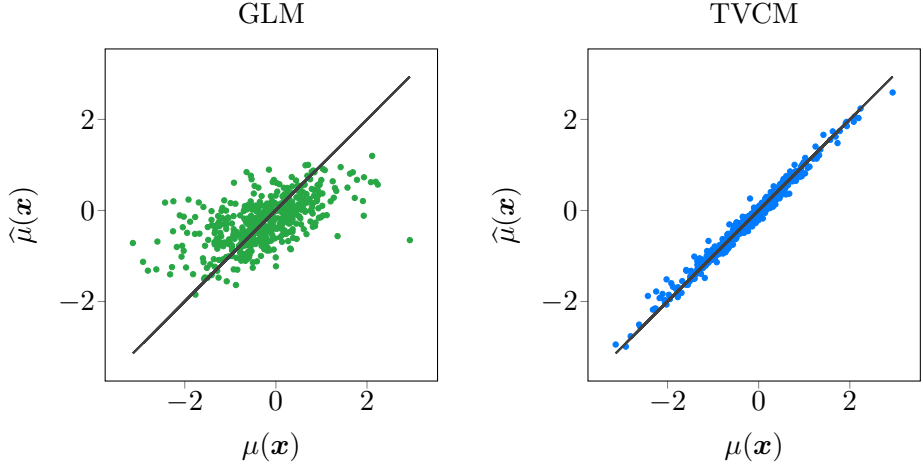

The estimated means $\widehat{\mu}(\bx)$ compared with the true model means $\mu(\bx)$
on $500$ randomly selected observations from the test set
for the GLM and TVCMs respectively are shown in
Figure~\ref{fig:simulated_data_estimated_means}.
It is apparent from Figure~\ref{fig:simulated_data_estimated_means} that the TVCM
model is able to capture the true regression function well, while the GLM
model is not.
These results are similar to those of the LocalGLMnet (see Figure 1 in~\cite{richman2023localglmnet}).

The number of parameters $\kappa_j$ from the early stopping procedure as
well as the constant parameters $\widehat{\beta}_j^{{\operatorname{GLM}}}$,
are shown in Table~\ref{tab:simulated_data_kappa}.
The table also presents th $\text{FI}^*_j$ from~\eqref{eq:variable_importance}
for the TVCM.
\begin{table}[ht]
    \centering
    \begin{tabular}{l|rrr}
{} & $\kappa_j$ & $\widehat{\beta}_{j}^{\operatorname{GLM}}$ & $\text{FI}^*_j$ \\
\midrule
\begin{filecontents*}{data/sim_parameters.csv}
featureName,kappaValue,betaValue,variableImportance
$x_1$,0,0.500,0.33
$x_2$,469,-0.001,0.04
$x_3$,563,0.033,0.24
$x_4$,271,0.008,0.13
$x_5$,363,-0.001,0.13
$x_6$,321,0.123,0.08
$x_7$,24,-0.001,0.00
$x_8$,220,-0.001,0.04
\end{filecontents*}
\csvreader[
 head to column names,
 late after line=\\
]{data/sim_parameters.csv}{}%
{\featureName & \kappaValue & \betaValue & \variableImportance}
\bottomrule
\end{tabular}

    \caption{Hyperparameters $\kappa_j$, GLM estimates
        $\widehat{\beta}_j^{(0)}$
        and $\text{FI}^*_j$ scores
        for the TVCM in the simulated
        data example.}
    \label{tab:simulated_data_kappa}
\end{table}
Coefficient function estimate $\widehat{\beta}_1$ got the expected number of estimators, $\kappa_1 = 0$,
since this function contains no $\bx$-dependent terms.
For $\widehat{\beta}_7$, the number of estimators $\kappa_7$ is non-zero even though
$\beta_7 \equiv 0$ in the true regression function, which means that
the model is overfitting to the training data.
The number of estimators is, however, not very large compared to the other
coefficient function estimates.
The fact that $\kappa_8$ is not zero is likely a consequence of the correlation
between $X_2$ and $X_8$.

As for the GLM terms $\widehat{\beta}_j^{\operatorname{GLM}}$, note that the expected value of
the coefficient functions, $\mathbb{E}[\beta_j(X)]$, is equal to $0$ for $j = 2,4,5,7,8$,
which is mirrored in the GLM terms $\widehat{\beta}_j^{\operatorname{GLM}}$ being close to zero.
Also, note that $\mathbb{E}[\beta_1(X)] = 0.5$ and $\mathbb{E}[\beta_6(X)] = 0.125$,
and that the GLM terms $\widehat{\beta}_1^{\operatorname{GLM}}$ and $\widehat{\beta}_6^{\operatorname{GLM}}$ are
close to these values.
For $\beta_3$, the GLM term $\widehat{\beta}_3^{\operatorname{GLM}}$ is close to zero, which
is not expected since $\mathbb{E}[\beta_3(X)] \approx 0.255$.

The $\mathrm{FI}_j^*$s from~\eqref{eq:variable_importance} for the TVCM in
Table~\ref{tab:simulated_data_kappa} is the highest for coefficient function estimate $\widehat{\beta}_1$,
which is expected since the constant $\beta_1$ is larger than
the other coefficient functions are on average.
The term $\widehat{\beta}_3$ gets the second highest $\mathrm{FI}_j^*$, which
is also expected since $\beta_3(\bx)$ has a large amplitude.
In Figure~\ref{fig:simulated_data_regression_attentions} the
true coefficient functions $\beta_j(\bx)$ are compared with their estimates based on the TVCM.

\begin{figure}[ht]
    \centering
\newcommand{\nfeatures}{8}

\newcommand{\feature}[1]{%
    \pgfmathparse{int(#1+1)}%
    $x_{\pgfmathprintnumber[assume math mode=true, fixed zerofill, precision=0]{\pgfmathresult}}$%
}

\newcommand{\betahat}[1]{%
    \pgfmathparse{int(#1+1)}%
    $\widehat{\beta}_{\pgfmathprintnumber[assume math mode=true]{\pgfmathresult}}$%
}

\begin{tikzpicture}
\begin{axis}[
    colorbar,
    colormap={greyToWarm}{color(0cm)=(LightGray); color(1cm)=(PleasantBlue)},
    xmin=-0.5, xmax=\nfeatures-0.5,
    ymin=-0.5, ymax=\nfeatures-0.5,
    axis equal image,
    xtick=data,
    ytick = data,
    xticklabel={\feature{\tick}},
    yticklabel={\betahat{\tick}},
    nodes near coords={\pgfmathprintnumber[fixed, precision=2]\pgfplotspointmeta},
    nodes near coords style={text = DarkGray, anchor=center, font=\footnotesize},
    y dir=reverse,
    axis line style = {draw=none},
]
\addplot[
    matrix plot*,
    point meta=explicit,
    mesh/rows=\nfeatures,
    mesh/cols=\nfeatures,
] table [col sep=comma, x=feature, y=beta, meta=value] {data/sim_feature_importance.csv};
\end{axis}
\end{tikzpicture}
    \caption{
        Feature importance scores for the simulated data example.
        Cell $(x_j,\widehat{\beta}_k)$ corresponds to the feature importance score of
        feature $x_j$ for the coefficient function $\widehat{\beta}_k$, i.e.\ the relative
        loss reduction from splits in the trees of coefficient function estimate $\widehat{\beta}_k$
        that used feature $x_j$.
        All rows are normalised to sum to one.
        Note that $\widehat{\beta}_0$ is the intercept and does not depend on any features,
        and that $\kappa_1 = 0$ means that coefficient function estimate $\widehat{\beta}_1$ consists
        of the GLM initiation only.
    }
    \label{fig:simulated_data_feature_importance}
\end{figure}
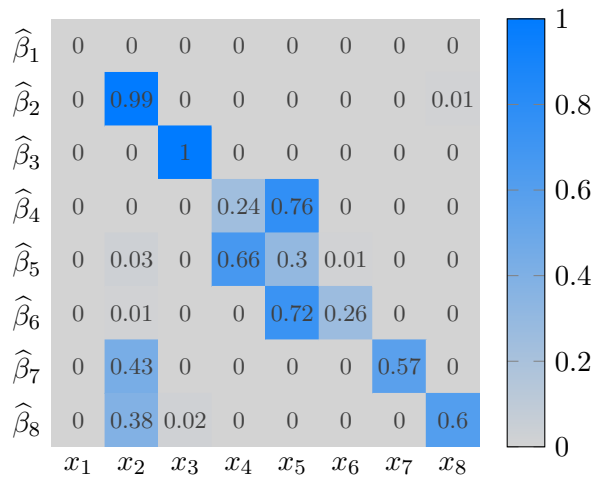

\begin{figure}[H]
    \centering
    \input{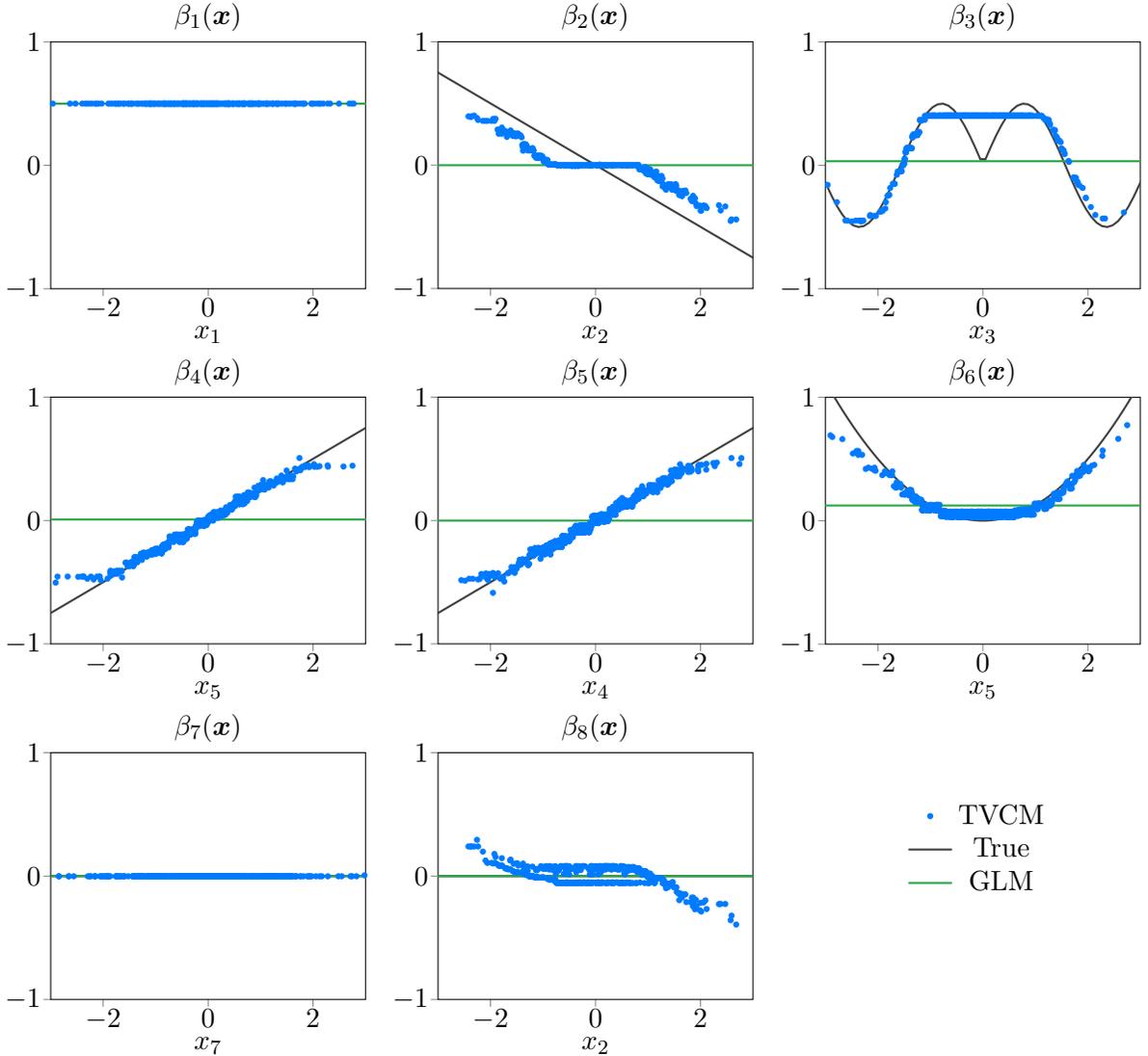}
\begin{tikzpicture}

    \begin{groupplot}[
        group style={
            group size=3 by 3,
            vertical sep=1.5cm,
        },
        width=0.38*\textwidth,
        height=0.32*\textwidth,
        xtick pos=left,
        ytick pos=left,
        tick style={major tick length=-1mm},
        xmin=-3, xmax=3,
        ymin=-1,ymax=1,
        title style = {yshift=-0.5em},
        xlabel style = {yshift=0.4em},
        ytick={-1,0,1},
    ]

    \nextgroupplot[
    xlabel = {$x_1$},
    title = {$\beta_1(\bx)$},
    ]
    \addplot[
        only marks,
        mark size=1pt,
        mark options={draw=PleasantBlue,fill=PleasantBlue,opacity=0.3},
    ] table[
        x=X1,
        y=beta_X1,
        col sep=comma,
    ] {data/sim_attentions.csv};
    \addplot[no marks,
        DarkGray,
        samples=2,
        thick
    ] {0.5};
    \addplot[FreshGreen, thick] {\simBetaI};

    \nextgroupplot[
    xlabel = {$x_2$},
    title = {$\beta_2(\bx)$},
    ]
    \addplot[
        only marks,
        mark size=1pt,
        mark options={draw=PleasantBlue,fill=PleasantBlue,opacity=0.3},
    ] table[
        x=X2,
        y=beta_X2,
        col sep=comma,
    ] {data/sim_attentions.csv};
    \addplot[no marks,
        DarkGray,
        samples=2,
        thick
    ] {-0.25*x};
    \addplot[FreshGreen, thick] {\simBetaII};

    \nextgroupplot[
    xlabel = {$x_3$},
    title = {$\beta_3(\bx)$},
    ]
    \addplot[
        only marks,
        mark size=1pt,
        mark options={draw=PleasantBlue,fill=PleasantBlue,opacity=0.3},
    ] table[
        x=X3,
        y=beta_X3,
        col sep=comma,
    ] {data/sim_attentions.csv};
    \addplot[no marks,
        DarkGray,
        samples=100,
        thick
    ] {0.5 * sign(x) * sin(deg(2*x))};
    \addplot[FreshGreen, thick] {\simBetaIII};

    \nextgroupplot[
    xlabel = {$x_5$},
    title = {$\beta_4(\bx)$},
    ]
    \addplot[
        only marks,
        mark size=1pt,
        mark options={draw=PleasantBlue,fill=PleasantBlue,opacity=0.3},
    ] table[
        x=X5,
        y=beta_X4,
        col sep=comma,
    ] {data/sim_attentions.csv};
    \addplot[no marks,
        DarkGray,
        samples=100,
        thick
    ] {0.25 * x};
    \addplot[FreshGreen, thick] {\simBetaIV};

    \nextgroupplot[
    xlabel = {$x_4$},
    title = {$\beta_5(\bx)$},
    ]
    \addplot[
        only marks,
        mark size=1pt,
        mark options={draw=PleasantBlue,fill=PleasantBlue,opacity=0.3},
    ] table[
        x=X4,
        y=beta_X5,
        col sep=comma,
    ] {data/sim_attentions.csv};
    \addplot[no marks,
        DarkGray,
        samples=100,
        thick
    ] {0.25 * x};
    \addplot[FreshGreen, thick] {\simBetaV};

    \nextgroupplot[
    xlabel = {$x_5$},
    title = {$\beta_6(\bx)$},
    ]
    \addplot[
        only marks,
        mark size=1pt,
        mark options={draw=PleasantBlue,fill=PleasantBlue,opacity=0.3},
    ] table[
        x=X5,
        y=beta_X6,
        col sep=comma,
    ] {data/sim_attentions.csv};
    \addplot[no marks,
        DarkGray,
        samples=100,
        thick
    ] {0.125 * x^2};
    \addplot[FreshGreen, thick] {\simBetaVI};

    \nextgroupplot[
    xlabel = {$x_7$},
    title = {$\beta_7(\bx)$},
    ]
    \addplot[
        only marks,
        mark size=1pt,
        mark options={draw=PleasantBlue,fill=PleasantBlue,opacity=0.3},
    ] table[
        x=X7,
        y=beta_X7,
        col sep=comma,
    ] {data/sim_attentions.csv};
    \addplot[no marks,
        DarkGray,
        samples=2,
        thick
    ] {0};
    \addplot[FreshGreen, thick] {\simBetaVII};

    \nextgroupplot[
    xlabel = {$x_2$},
    title = {$\beta_8(\bx)$},
    legend style = {
    draw = none,
    xshift = 0.275*\textwidth,
    yshift = -0.5cm
    }
    ]
    \addplot[
        only marks,
        mark size=1pt,
        mark options={draw=PleasantBlue,fill=PleasantBlue,opacity=0.3},
    ] table[
        x=X2,
        y=beta_X8,
        col sep=comma,
    ] {data/sim_attentions.csv};
    \addlegendentry{TVCM}
    \addplot[no marks,
        DarkGray,
        samples=2,
        thick
    ] {0};
    \addlegendentry{True}
    \addplot[FreshGreen, thick] {\simBetaVIII};
    \addlegendentry{GLM}

    \end{groupplot}

\end{tikzpicture}
    \caption{
        Coefficient function estimates $\widehat{\beta}_j(\bx)$ for the TVCM
        estimates on the simulated data set
        for different values of $x_l$, where $l$ is the effect modifier with the highest
        feature importance score for each coefficient function.
        The GLM initiation $\widehat{\beta}_j^{(0)}$ is also shown, as well as the
        true regression definitions from Table~\ref{tab:simulated_true_regression_attentions}.}
    \label{fig:simulated_data_regression_attentions}
\end{figure}

The feature importance score $\text{FI}_j$, as defined by~\cite{friedman2001greedy}, of each effect modifier
feature on each individual coefficient function for the TVCM are shown in
Table~\ref{fig:simulated_data_feature_importance}.
The feature importances in Figure~\ref{fig:simulated_data_feature_importance}
 are mainly in line with expectations, with the important features for
each coefficient function estimate being the ones specified in Table~\ref{tab:simulated_true_regression_attentions}.
For $\beta_4$ and $\beta_5$, the coefficient function estimates however also seem to use their respective
features $x_4$ and $x_5$ to model the interaction term $x_4x_5$.
Interestingly, $\widehat{\beta}_5$ does not use $x_6$ at all, which means that the interaction term
$x_5^2x_6$ is modeled solely in $\widehat{\beta}_6$.
Also, $\widehat{\beta}_8$ is mostly using $x_2$ as a feature, which indicates that it is using the correlation
between $x_2$ and $x_8$ to model the $x_2^2$ term.
The feature interaction analysis in~\cite{richman2023localglmnet} indicate largely the same results
(see Figure 4 in~\cite{richman2023localglmnet} and the corresponding discussion).

The estimated coefficient functions $\widehat{\beta}_j(\bx)$ from the TVCM are shown in
Figure~\ref{fig:simulated_data_regression_attentions}.
The function estimates are shown as compared to the feature values of the effect modifier in the
true regression function in Table~\ref{tab:simulated_true_regression_attentions}.
The coefficient function estimates from the TVCM are compared with the corresponding $\beta_j(\bx)$s from
Table~\ref{tab:simulated_true_regression_attentions}.
For $\widehat{\beta}_1$, $\widehat{\beta}_7$ and $\widehat{\beta}_8$, the $x$-axis uses the feature values of the effect modifier
with the highest feature importance score from Table~\ref{fig:simulated_data_feature_importance}, since
the true coefficient functions are not feature-dependent for these coefficient functions.
From Table~\ref{tab:simulated_data_mse} and Figure~\ref{fig:simulated_data_regression_attentions}
it is seen that the TVCM is able to capture the true coefficient functions
well, and that the terms specified by Table~\ref{tab:simulated_true_regression_attentions}
also are the ones that the TVCM is modeling. 
This implies that the identification problem discussed in Remark~\ref{remark:regression_attentions} is not material.
Note that the coefficient function estimates $\widehat{\beta}_j$ for $j=2,3$ is not
very accurate for observations close to $x_j=0$.
This is not surprising, since the term $\beta_j(\bx)x_j$ is small in those observations
and does, hence, not affect the mean $\mu(\bx)$ much.
Also, the shape of the coefficient function estimate $\widehat{\beta}_8$ indicates that this is affected by the
correlation between $x_2$ and $x_8$.
While Figure~\ref{fig:simulated_data_regression_attentions} is not identical to Figure 2 in~\cite{richman2023localglmnet},
which is used for feature selection, it shows similar results for the estimated regression attentions.
For coefficient function estimates $\widehat{\beta}_1$, $\widehat{\beta}_2$, $\widehat{\beta}_3$,
and $\widehat{\beta}_7$, both models are able to capture the true coefficient functions well.
For $\beta_1$ and $\beta_7$, the TVCM is better at capturing the
flat structure of the coefficient, where the LocalGLMnet model is more noisy and seems to be slightly
overfitting.
For $\beta_8$, the LocalGLMnet coefficient function estimates are more noisy, but of the same magnitude
as the TVCM equivalents.

        \subsection{Real data}
        \label{subsec:real_data}
        In order to be able to compare with the results for the LocalGLMnet model
from~\cite{richman2023localglmnet}, the real world example is based on
the \texttt{FreMTPLfreq2} dataset from~\cite{CASdatasets}.
The dataset is a widely used non-life insurance dataset with the number and severity
of claims for a portfolio of French motor third-party liability insurance policies 
(see~\cite[Chapter 13.1]{wuthrich2023statistical} for a detailed description).

As in~\cite{richman2023localglmnet}, the focus will be on claim count modelling using
Poisson regression models with duration weights. 
That is, it is assumed that
\[
	Y \mid \bX = \bx_i, W = w_i \sim \operatorname{Pois}(w_i \mu(\bx_i)).
\]
where $Y$ is the number of claims, $\bX$ is a vector of policy data, and $W$ is the
duration of exposure.
By considering the number of claims as the response $Y_i$, the duration of exposure
as weights $w_i$, and policy data $\bx_i$, the dataset can be used to fit
the TVCM, using the Poisson deviance as the loss function.
Note that this requires that the duration weights be incorporated in the
loss function, which is a simple extension of the algorithms described in this
paper.

The dataset contains $678\,008$ observations, where each observation corresponds to
a policy.
Further, the data cleaning and train-test-split from~\cite{wuthrich2023statistical}
will be used.
This results in a dataset $(y_i,w_i,\bx_i)_i$ which is split into a training dataset
of size $610\,206$ and a test dataset of size $67\,801$.
The feature vector $\bx_i$ contains $6$ ordinal features and $3$ categorical features.
The categorical features (\texttt{Region}, \texttt{VehGas} and \texttt{VehBrand}) are
handled using one-hot encoding in line with the discussion in
Section~\ref{subsec:categorical_features}.
The one-hot encoding of the categorical features is done for the effect modifier
features as well.
This is due to limitations in the \texttt{python} implementation, which relies on
the \texttt{DecisionTreeRegressor} from \texttt{sklearn}, which does not support
categorical features.

The TVCM is tuned using the early stopping algorithm given by Algorithm 3
in~\cite{delong2023note} using two equally sized training-validation splits.
The other tree and boosting hyperparameters are set to reasonable values without
any in-depth evaluation; the max depth is set
to $2$, the minimum samples per leaf node is set to $20$, and the learning rate
$\epsilon_j$ is set to $0.01$ for all $j$.

The loss results in terms of Poisson deviance for the training and test data are
summarised in Table~\ref{tab:real_data_loss}.
\begin{table}[ht]
    \centering
    \begin{tabular}{lrr}
{} & Train & Test \\
\midrule
\begin{filecontents*}{data/real_loss.csv}
Model,Train,Test
Intercept,25.213,25.445
GLM,24.177,24.217
GBM,23.846,23.890
TVCM,23.751,23.839
LocalGLMnet,23.728,23.945
\end{filecontents*}
\csvreader[
 head to column names,
 late after line=\\
]{data/real_loss.csv}{}%
{\Model & \Train & \Test}
\bottomrule
\end{tabular}

    \caption{Poisson deviance results ($10^{-2}$) for the real data example.}
    \label{tab:real_data_loss}
\end{table}
As can be seen, the TVCM achieves a lower out-of-sample Poisson
deviance than both the standard GBM and the LocalGLMnet model
from~\cite{richman2023localglmnet}.
While the deviance score differences are small, this shows that the training
algorithm is able to achieve a fit that is comparable to its peers.
It is also apparent that the TVCM is able to achieve a lower
out-of-sample deviance than the GLM, which indicates that there is structure
in the data beyond the linear term structure of a GLM.

\begin{figure}[ht]
    \centering
    \begin{tikzpicture}
\begin{axis}[
    legend cell align={left},
    legend style={
        fill opacity=0.8,
        draw opacity=1,
        draw=none,
        text opacity=1,
        at={(0.03,0.97)},
        anchor=north west,
    },
    width = 10cm,
    height = 6cm,
    tick align=outside,
    tick pos=left,
    x grid style={darkgray176},
    xlabel={Observation (ordered)},
    xmin=-2296, xmax=71052,
    xtick style={color=black},
    y grid style={darkgray176},
    ylabel={Claim count},
    ymin=0, ymax=0.5,
    scaled x ticks = false,
    xtick = {0, 30000, 60000},
    xticklabels = {0, $30{,}000$, $60{,}000$},
]
\addplot[DarkGray,thick] table [x=index, y=y, col sep=comma] {data/real_mu_hat.csv};
\addlegendentry{y}

\addplot[BrightRed,thick] table [x=index, y=Intercept, col sep=comma] {data/real_mu_hat.csv};
\addlegendentry{Intercept}

\addplot[FreshGreen,thick] table [x=index, y=GLM, col sep=comma] {data/real_mu_hat.csv};
\addlegendentry{GLM}

\addplot[WarmOrange,thick] table [x=index, y=GBM, col sep=comma] {data/real_mu_hat.csv};
\addlegendentry{GBM}

\addplot[PleasantBlue,thick] table [x=index, y=TVCM, col sep=comma] {data/real_mu_hat.csv};
\addlegendentry{TVCM}

\end{axis}
\end{tikzpicture}
    \caption{
        Out of sample predictions of $\mu(\bx)$ on the real data set,
        where $\mu(\bx) = \mathbb{E}[Y_i \mid \bX = \bx_i, W = w_i]/w_i$.
        The predictions have been ordered according to the $\mu(\bx)$-predictions
        from the TVCM and averaged using a rolling mean of $1\,000$.
        For the true observations $y$, the rolling mean is taken over the true
        number of claims and divided by the rolling mean of the duration of exposure.
    }
    \label{fig:real_data_predictions}
\end{figure}
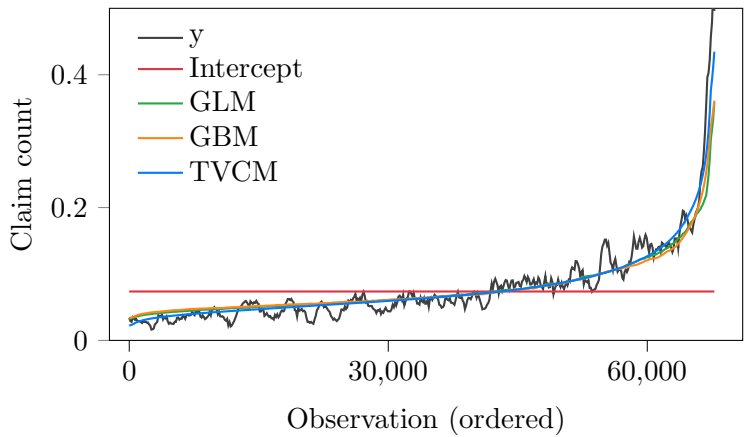

Figure~\ref{fig:real_data_predictions} shows the outcome on the test set compared to model predictions.
The LocalGLMnet model from~\cite{richman2023localglmnet} is not included in this
figure since we do not have access to the individual predictions from this model.
As can be seen in Figure~\ref{fig:real_data_predictions}, the TVCM
captures more of the structure in the data than the GLM and the GBM, especially for
more extreme values.
For the more moderate values, the three models more or less coincide.

The number of estimators $\kappa_j$ for the different features can be seen in
Table~\ref{tab:real_data_estimators}, together with the constant
terms for every coefficient function, $\widehat{\beta}_j^{\operatorname{GLM}}$.
This table also presents the $\text{FI}^*_j$ scores
as defined in~\eqref{eq:variable_importance}.
Note that the $\text{FI}^*_j$ scores are not calculated for the categorical
features since they are not on the same scale as the other features,
see~\cite{richman2023localglmnet} for a discussion.
\begin{table}[ht]
    \centering
    \begin{tabular}{l|rrr}
{} & $\kappa_j$ & $\widehat{\beta}_{j}^{\operatorname{GLM}}$ & $\text{FI}^*_j$ \\
\midrule
\begin{filecontents*}{data/real_parameters.csv}
featureName,kappaValue,betaValue,variableImportance
VehPower,33,0.07,0.08
VehAge,46,-0.10,0.12
Density,18,-0.01,0.01
DrivAge,241,0.08,0.11
BonusMalus,59,0.41,0.49
Area,91,0.15,0.18
VehGas,(139{,}187),(-0.70{,}-0.87),-
VehBrand,(6{,}145),(-0.05{,}0.01),-
Region,(0{,}187),(-0.01{,}0.25),-
\end{filecontents*}
\csvreader[
 head to column names,
 late after line=\\
]{data/real_parameters.csv}{}%
{\featureName & \kappaValue & \betaValue & \variableImportance}
\bottomrule
\end{tabular}

    \caption{Number of estimators $\kappa_j$ and GLM terms
        $\widehat{\beta}_{j}^{(0)}$ for the real
    data example.
    For the categorical features (\texttt{VehGas}, \texttt{VehBrand}, and
    \texttt{Region}), intervals
    of $\kappa_j$ and $\beta_{j0}$ are presented.
    }
    \label{tab:real_data_estimators}
\end{table}
From Table~\ref{tab:real_data_estimators}, it is seen that most features are used in the coefficient functions,
except for parts of the one-hot encoded regions in \texttt{Region}.
The $\text{FI}^*_j$ scores of~\eqref{eq:variable_importance} for the
different features indicate that the \texttt{BonusMalus} feature
gets the largest absolute values of its coefficient function estimates,
which agrees with~\cite{richman2023localglmnet}.
However, the feature with the second highest $\text{FI}^*_j$ according to the TVCM is \texttt{Area}, which
contradicts the findings in~\cite{richman2023localglmnet}, whose analysis suggests that the feature should be dropped.
This could be because \texttt{Area} is highly correlated with \texttt{Density}, which is given a higher
value of $\text{FI}^*_j$ in~\cite{richman2023localglmnet}.

\begin{figure}[ht]
    \centering
    \newcommand{\nfeatures}{9}

\begin{tikzpicture}
\begin{axis}[
    colorbar,
    colormap={greyToWarm}{color(0cm)=(LightGray); color(1cm)=(PleasantBlue)},
    xmin=-0.5, xmax=\nfeatures-0.5,
    ymin=-0.5, ymax=\nfeatures-0.5,
    axis equal image,
    xtick=data,
    ytick=data,
    xticklabels={
        {\texttt{VehPower}},
        {\texttt{VehAge}},
        {\texttt{Density}},
        {\texttt{DrivAge}},
        {\texttt{BonusMalus}},
        {\texttt{Area}},
        {\texttt{VehGas}},
        {\texttt{VehBrand}},
        {\texttt{Region}}
    },
    xticklabel style={rotate=45, anchor=north east, inner sep=0.5mm},
    yticklabels={
        {$\widehat{\beta}_{\mathtt{VehPower}}$},
        {$\widehat{\beta}_{\mathtt{VehAge}}$},
        {$\widehat{\beta}_{\mathtt{Density}}$},
        {$\widehat{\beta}_{\mathtt{DrivAge}}$},
        {$\widehat{\beta}_{\mathtt{BonusMalus}}$},
        {$\widehat{\beta}_{\mathtt{Area}}$},
        {$\widehat{\beta}_{\mathtt{VehGas}}$},
        {$\widehat{\beta}_{\mathtt{VehBrand}}$},
        {$\widehat{\beta}_{\mathtt{Region}}$}
    },
    nodes near coords={\pgfmathprintnumber[fixed, precision=2]\pgfplotspointmeta},
    nodes near coords style={text=DarkGray, anchor=center, font=\scriptsize},
    y dir=reverse,
    axis line style = {draw=none},
]
\addplot[
    matrix plot*,
    point meta=explicit,
    mesh/rows=\nfeatures,
    mesh/cols=\nfeatures,
] table [col sep=comma, x=feature, y=beta, meta=value] {data/real_feature_importance.csv};
\end{axis}
\end{tikzpicture}
    \caption{Feature importance scores for the real data example.
    Cell $(j,\widehat{\beta}_k)$ corresponds to the feature importance score of
    feature $j$ for the coefficient function estimate
    $\widehat{\beta}_k$, i.e.\ the relative
    loss reduction from splits in the trees of 
    coefficient function estimate $\widehat{\beta}_k$ that used feature $j$.
    All rows are normalised to sum to one.
    For the categorical features (\texttt{VehGas}, \texttt{VehBrand}, and
    \texttt{Region}), the feature importance scores have been summed over the
    different categories before normalisation.
    }
    \label{fig:real_data_feature_importances}
\end{figure}
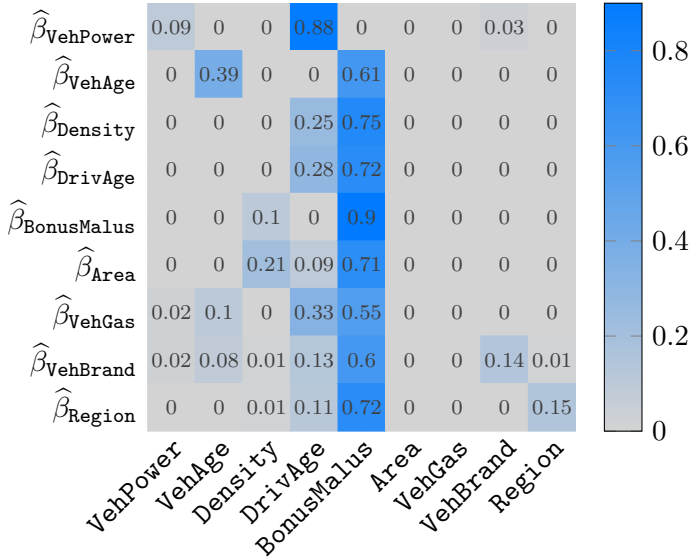

The feature importance score $\text{FI}_j$, as defined by~\cite{friedman2001greedy}, of each effect modifier feature on each individual
coefficient function for the TVCM can be seen in
Figure~\ref{fig:real_data_feature_importances}.
Similarly to the findings in~\cite{richman2023localglmnet}, see their Figure 8,
\texttt{BonusMalus} is the most influential feature on the coefficient function estimates
as a whole.
Both the TVCM and LocalGLMnet models agree that \texttt{DrivAge} is the second most
influential feature for most coefficient function estimates.
The fact that many feature importance scores in Figure~\ref{fig:real_data_feature_importances} 
are close to zero suggests that the regression models can be reduced by removing features from the modeling. 
This, however, should be backed by, e.g., a bootstrap analysis.

\begin{figure}[ht]
    \centering
    \input{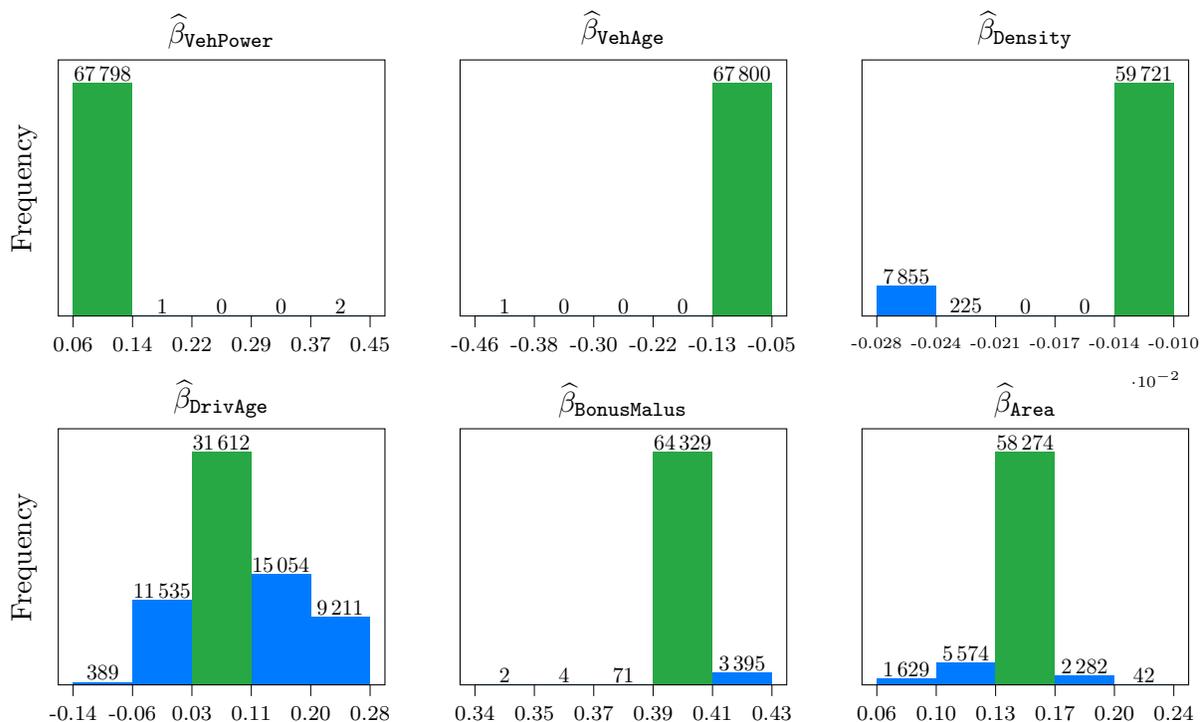}
    \caption{
        Histograms of the coefficient function estimates $\widehat{\beta}_j(\bx)$
        for the TVCM estimates on the test set of the real data set for the different
        continuous predictive features $x_j$.
        The bin containing the constant term $\widehat{\beta}_j^{(0)}$ is coloured
        in green.
    }
    \label{fig:real_data_regression_attentions}
\end{figure}

A histogram of the coefficient function estimates $\widehat{\beta}_j(\bx)$
for the continuous predictive features $x_j$ is shown in
Figure~\ref{fig:real_data_regression_attentions}.
As can be seen, most of the coefficient function estimates do not vary much from the
constant term $\widehat{\beta}_j^{(0)}$.
This differs from the corresponding Figure~5 in~\cite{richman2023localglmnet},
where the coefficient function estimates vary more dramatically.
Further, the TVCM seems to produce coefficient function estimates that are
closer to the GLM than the LocalGLMnet model from~\cite{richman2023localglmnet}.
This is likely a consequence of the fact that the local GLM formulation comes
with a native identification problem -- there are likely many different
local GLMs that provide drastically different coefficient function estimates
but still produce similar predictions -- a hypothesis that is
supported by the loss results in Table~\ref{tab:real_data_loss}.
Without access to the individual predictions from the LocalGLMnet model
from~\cite{richman2023localglmnet}, it is not possible to analyse this
hypothesis further.

    \section{Conclusion}
    \label{sec:conclusion}
    This paper introduces an interpretable tree-based VCM built using cyclic GBMs,
following the ideas as those that underpin the neural network based LocalGLMnet
from~\cite{richman2023localglmnet}, in an approach that is similar to that of the
boosted VCM from~\cite{zhou2022decision}.
The cyclic GBM from~\cite{delong2023note}
allows for dimension-wise early stopping in the training of the
regression coefficient functions, as well as the use of
coefficient function-wise feature importance, which makes model tuning and
interpretation easier than for the models in~\cite{richman2023localglmnet} and~\cite{zhou2022decision}.
By using the dimension-wise feature importance scores it is also easy to identify potentially redundant features that could
be removed from the modeling of the regression coefficient functions. This procedure is simpler than the corresponding model 
reduction techniques discussed for the LocalGLMnet in~\cite{richman2023localglmnet}.
In~\cite{richman2023localglmnet} splines are used to assess interaction effects, whereas the current analyses rely on feature
importance scores.
By using splines it is possible to obtain more granular information on \textit{within}-feature dynamics;
something which is not possible when using feature importance scores.
There is, however, nothing that prevents us from analysing tree-based VCMs using splines.

The results of the tree-based VCM on simulated and real data shows that it outperforms
GLM and GBM models for the same regression problems, and produces results in terms of out of sample
loss that are comparable to those of the LocalGLMnet model.
Further, additional analyses indicates that the resulting tree-based VCM produces a model structure that is similar
to that of the LocalGLMnet model.

Concerning extensions, in~\cite{wang2014boosted} the use of elastic nets are discussed, which is also possible to use for the
current tree-based VCM.

    \section*{Acknowledgements}
    \label{sec:acknowledgements}
    The authors would like to thank Professor Mario V.~W{\"u}thrich for comments on an early draft as well as for providing the data from~\cite{richman2023localglmnet}.
The authors would also like to thank Ronald Richman for discussions on the topic of this paper on an early stage.
M.~Lindholm acknowledges financial support from Stiftelsen Länsförsäkringsgruppens Forsknings- och Utvecklingsfond, grant P9/20 ``Machine learning methods in non-life insurance''.
The \texttt{python} implementation of the special case of the VCM presented in Section~\ref{sec:examples} is available at
\url{https://github.com/henningzakrisson/local-glm-boost} under the name \texttt{LocalGLMboost}.

    \bibliography{references}
    
    \appendix
    \section{Comments on regression coefficient functions in VCMs}
    \label{app:l2_calculations}

	\subsection{Interpretation of regression coefficient functions in VCMs}
	\label{app:12_coefficient_function_interpretations}
	Concerning the interpretation of the coefficient functions, consider the situation with an $L^2$-loss with identity link-function and assume that you know the correct $\bz_j$s, $j = 1, \ldots, p$; here neglecting the constant intercept term. 
It then holds that
\begin{align}\label{eq: regression attentions L2}
	\beta_j^*(\bZ_j) &= \frac{\operatorname{E}[X_j (Y - \sum_{k \neq j}\beta_k^*(\bZ_k)X_k) \mid \bZ_j]}{\operatorname{E}[X_j^2 \mid \bZ_j]} \nonumber\\
		&= \operatorname{E}\left[\frac{X_j^2}{\operatorname{E}[X_j^2 \mid \bZ_j ]} \frac{Y - \sum_{k \neq j}\beta_k^*(\bZ_k)X_k}{X_j} ~\bigg|~ \bZ_j\right]
\end{align}
minimises
\[
	\operatorname{E}[(Y - \sum_{j = 1}^p \beta_k^*(\bZ_k)X_k)^2],
\]
see (6) in~\cite{hastie1993varying} together with the surrounding discussion.
Note that we use the notation $\beta_j^*$ to stress that we assume an infinite amount 
of data, since \eqref{eq: regression attentions L2} is based on population quantities.
Not surprisingly, from \eqref{eq: regression attentions L2} it is seen that the coefficient functions need to be estimated jointly.
Moreover, from  \eqref{eq: regression attentions L2} it is seen that the $\beta_j^*(\bZ_j)$s corresponds to weighted conditional expectations, 
with weights proportional to $X_j^2$. In particular, note that each $\beta_j^*(\bZ_j)$ corresponds to the remaining conditional expectation 
to be explained after having adjusted for the effect of the coefficient functions $\beta_k^*(\bZ_k), k \neq j$. 

Further, note that if the true mean function $\mu(\bX) = \operatorname{E}[Y \mid \bX]$ is given by
\begin{align*}
	\mu(\bX) := \sum_{j = 1}^p \gamma_j(\bZ_j)X_j,
\end{align*}
it directly follows from \eqref{eq: regression attentions L2} that
\begin{align}\label{eq: true attentions}
	\beta_j^*(\bZ_j) = \gamma_j(\bZ_j) + \operatorname{E}\left[\frac{X_j^2}{\operatorname{E}[X_j^2 \mid \bZ_j ]}\frac{\sum_{k \neq j}(\gamma_k(\bZ_k) - \beta_k^*(\bZ_k))X_k }{X_j} ~\bigg|~ \bZ_j\right].
\end{align}
Thus, from \eqref{eq: true attentions} it follows that $\beta_j^*(\bZ_j)$ is, a possibly noisy, estimator of $\gamma_j(\bZ_j)$;
see also the discussion in Section~\ref{sec:examples}.
This also connects to identifiability issues discussed in Remark~\ref{remark:regression_attentions}.
For the special case where $\gamma_j(\bZ_j) \equiv \gamma_j$, it follows that \eqref{eq: true attentions} reduces to precisely $\beta_j^* = \gamma_j$, see the separate calculations in Appendix~\ref{app:l2_linear_model_special_case}.

As a final note, the above $L^2$-discussion can be directly generalised to the Tweedie family, by taking conditional expected values of the corresponding estimating equations, see e.g.\ (3.39) in~\cite{ohlsson2010non}.

    \subsection{Details on $L^2$ linear model special case of VCM}
    \label{app:l2_linear_model_special_case}
    In the special case with a linear model we have that
\[
	Y = \sum_{j = 1}^p \gamma_j X_j + \epsilon,
\]
where $\epsilon$ is a random variable with mean 0 and variance $\sigma^2$.
This means that if we know that $\beta_j^*(\bZ_j) = \beta_j^*$ the conditional expectations in \eqref{eq: true attentions} 
are turned into unconditional expectations.
That is, \eqref{eq: true attentions} can be simplified to
\begin{align*}
	\beta_j^* &= \gamma_j + \sum_{k \neq j} (\gamma_k - \beta_k^*)\frac{\operatorname{E}[X_j X_k]}{\operatorname{E}[X_j^2]}\\
		&= \gamma_j + \sum_{k \neq j} (\gamma_k - \beta_k^*)c_{j, k},
\end{align*}
which can be rephrased according to
\begin{align*}
	\beta_j^* &= \gamma_j + (\boldsymbol\gamma - \bbeta^*)^\top \boldsymbol c_j,
\end{align*}
where $(\boldsymbol c_j)_k = c_{j, k}$ for all $k \neq j$ and $(\boldsymbol c_j)_j = 0$. Further, by introducing the 
$p \times p$-matrix $\boldsymbol C$ with columns given by $\boldsymbol c_j$ it follows that
\[
	(\bbeta^*)^\top = \boldsymbol\gamma^\top + (\boldsymbol\gamma - \bbeta^*)^\top \boldsymbol C,
\]
which can be re-written according to
\[
	(\bbeta^*)^\top (\boldsymbol I + \boldsymbol C) = \boldsymbol\gamma^\top (\boldsymbol I + \boldsymbol C),
\]
where $\boldsymbol I$ is a $p \times p$ identity matrix. Thus, when knowing the form of the $\bZ_j$s the relations given 
by~\eqref{eq: true attentions} provides the solution $\bbeta^* = \boldsymbol\gamma$,
 without any additional constraints on the $\bX_j$s or $\bZ_j$s, as it should.

    \subsection{Explicit $L^2$-calculations for the simulated example in Section~\ref{subsec:simulated_data}}
    \label{app:l2_numerical_example}
    We will now analyse the simulated Gaussian example from Section~\ref{subsec:simulated_data} based on the discussion in 
Section~\ref{subsec:local_glm} on $L^2$-losses that build on~\cite{hastie1993varying}.
The aim is to identify the correct basis of comparison for the estimated 
coefficient functions for the situation where the
true $\mu(\bx)$-function is known and satisfies the local GLM, or varying-coefficient regression formulation, 
from \eqref{eq:local_glm_structure} in Section~\ref{subsec:local_glm}.

From Table~\ref{tab:simulated_true_regression_attentions} it follows that 
$\bZ_1 = \bZ_7 = \bZ_8 = \{\emptyset\}$,
$\bZ_2 = X_2$, $\bZ_3 = X_3$, 
$\bZ_4 = X_5$, $\bZ_5 = X_4$, and 
$\bZ_6 = X_5$.
Further, from the construction of the example in Section~\ref{subsec:simulated_data} it follows that all $X_j$s are independent except
for $X_2$ and $X_8$. Combining all of the above results in that \eqref{eq: true attentions} reduces to
\begin{align*}
	\beta_1^*(\bZ_1) &= \beta_1,\\
	\beta_2^*(\bZ_2) &= \beta_2(X_2) + c_2 / X_2,\\
	\beta_3^*(\bZ_3) &= \beta_3(X_3) + c_3 / X_3,\\
	\beta_4^*(\bZ_4) &= \beta_4(X_5) + c_4 X_5,\\
	\beta_5^*(\bZ_5) &= \beta_5(X_4) + c_5 X_4,\\
	\beta_6^*(\bZ_6) &= \beta_6(X_5) + c_6 X_6,\\
	\beta_7^*(\bZ_7) &= \beta_7,\\
	\beta_8^*(\bZ_8) &= \beta_8,
\end{align*}
where $c_j$ are constants, where a perfect fit corresponds to that all $c_j = 0$.
Consequently, a perfect fit when assuming perfect a priori knowledge of the $\bZ_j$s and the functional 
form of $\mu(\bX)$ agrees with the local GLM, or varying-coefficient regression structure, from \eqref{eq:local_glm_structure},
retrieves the original coefficient functions. That is, ideally $\beta_j^*(\bZ_j) = \beta_j(\bZ_j)$, 
$j = 1, \ldots, p$. This motivates using the $\beta_j(\bZ_j)$s from Table~\ref{tab:simulated_true_regression_attentions}
as a basis for comparison with the corresponding coefficient functions from the cyclically boosted VCM in 
Section~\ref{subsec:simulated_data}.

\end{document}